\documentclass[10pt,twocolumn,letterpaper]{article}

\usepackage[pagenumbers]{iccv} 

\usepackage{amssymb}
\usepackage{tcolorbox}
\usepackage{graphicx}
\usepackage{multirow}
\usepackage{makecell}
\usepackage{array}
\usepackage{booktabs}

\usepackage[accsupp]{axessibility}  

\definecolor{iccvblue}{rgb}{0.21,0.49,0.74}
\usepackage[pagebackref,breaklinks,colorlinks,allcolors=iccvblue]{hyperref}

\newcommand{\name}{LMM$_{\text{HiMTok}}$}

\def\paperID{4156} 
\def\confName{ICCV}
\def\confYear{2025}

\title{HiMTok: Learning Hierarchical Mask Tokens for Image \\Segmentation with Large Multimodal Model}

\author{
    Tao Wang$^*$$^1$, Changxu Cheng$^*$$^{\dagger}$$^1$, Lingfeng Wang$^*$$^2$, Senda Chen$^*$$^3$, Wuyue Zhao$^1$\\
    $^1$Uni-Ubi \quad $^2$Zhejiang University \quad $^3$Tongji University\\
    {\tt\small \{wangtaomarvel,ccx0127,sendachen586\}@gmail.com,}
    {\tt\small yayafengzi@zju.edu.cn,}
    {\tt\small zhaohongyi@uni-ubi.com}
}

\begin{document}

\twocolumn[{
    \renewcommand\twocolumn[1][]{#1}%
    \maketitle
    \vspace{-5mm}
    \centering
    \includegraphics[width=.98\linewidth]{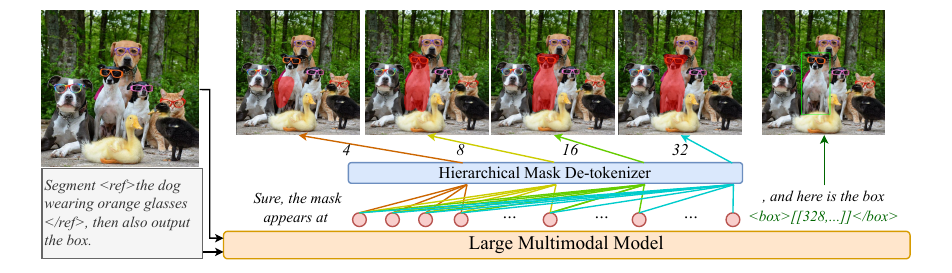}
    \vspace{-4mm}
    \captionof{figure}{The Large Multimodal Model integrated with HiMTok can progressively generate masks from coarse to fine, enhancing visual grounding capabilities. In the figure, mask tokens of varying lengths (4, 8, 16, and 32) represent different levels of granularity.}
    \vspace{-2mm}
    \label{fig:cover}
    \vspace{0.6cm}
}]

\renewcommand{\thefootnote}{\fnsymbol{footnote}}
\footnotetext{$^*$Equal contributions.}
\footnotetext{$^{\dagger}$Corresponding author.}

\begin{abstract}
    \indent The remarkable performance of large multimodal models (LMMs) has attracted significant interest from the image segmentation community.
    To align with the next-token-prediction paradigm, current LMM-driven segmentation methods either use object boundary points to represent masks or introduce special segmentation tokens, whose hidden states are decoded by a segmentation model requiring the original image as input.
    However, these approaches often suffer from inadequate mask representation and complex architectures, limiting the potential of LMMs.
    In this work, we propose the \textbf{Hi}erarchical \textbf{M}ask \textbf{Tok}enizer (HiMTok), which represents segmentation masks with up to 32 tokens and eliminates the need for the original image during mask de-tokenization.
    HiMTok allows for compact and coarse-to-fine mask representations, aligning well with the LLM next-token-prediction paradigm and facilitating the direct acquisition of segmentation capabilities.
    We develop a 3-stage training recipe for progressive learning of segmentation and visual capabilities, featuring a hierarchical mask loss for effective coarse-to-fine learning.
    Additionally, we enable bidirectional information flow, allowing conversion between bounding boxes and mask tokens to fully leverage multi-task training potential.
    Extensive experiments demonstrate that our method achieves state-of-the-art performance across various segmentation tasks,while also enhancing visual grounding and maintaining overall visual understanding.
    The codes are available at \url{https://github.com/yayafengzi/LMM-HiMTok}.
    \end{abstract}    
\section{Introduction}
\label{sec:intro}
With the rapid development of large multimodal models (LMMs), visual capabilities are advancing toward more generalized forms~\cite{liu2023llava,liu2024llava1_5,bai2023qwenvl,chen2024internvl,wu2024deepseekvl2,dubey2024llama3,yao2024minicpm}.
Image segmentation, a fundamental task in computer vision, has recently improved in terms of generalization and instruction-following ability through integration with LMMs~\cite{lai2024lisa,ren2024pixellm,zhang2024omg,zhang2024psalm}.
Since large language models (LLMs) were originally designed to generate only text tokens, developing an appropriate representation for image segmentation is both critical and challenging.

\begin{figure}[t]
    \centering
    \includegraphics[width=.95\linewidth]{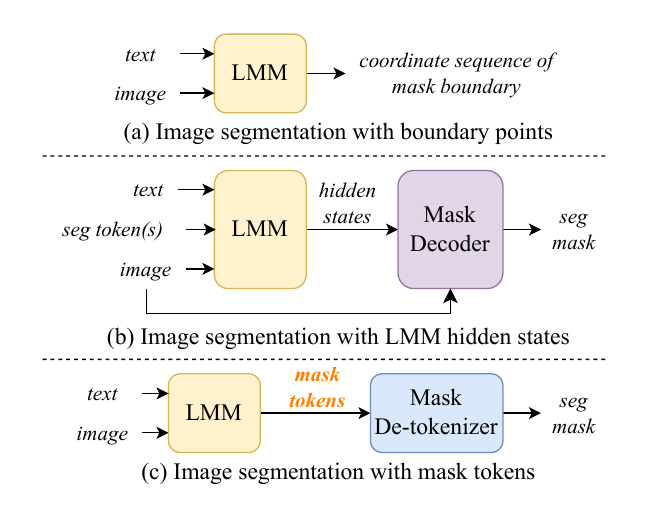}
    \caption{
        The three paradigms for LLM-based image segmentation.
        (a) Masks are represented as point sequences of boundaries. LMM output the coordinates directly.
        (b) LMM acts as a soft prompt generator for the image-conditioned mask decoder.
        (c) Ours. The LMM generates discrete mask tokens in the same manner as it generates text, and the mask de-tokenizer then converts these tokens into masks. This straightforward yet effective approach is enabled by our HiMTok.
    }
    \label{fig:intro_cmp}
    \vspace{-4mm}
\end{figure}

There are several paradigms for LLM-based image segmentation.
Some works represent a segmentation mask as a sequence of boundary points, which can be learned by a sequence-to-sequence framework~\cite{zhu2024llafs,wang2023visionllm,pramanick2024jack,chen2022unified,zhu2022seqtr}, as shown in \cref{fig:intro_cmp} (a). 
However, using \textbf{a limited number of polygon vertices} can impede accuracy, particularly in representing masks of complex shapes or multiple regions~\cite{pramanick2024jack,zhu2024llafs}.
Some other LMM-based segmentation methods~\cite{lai2024lisa,ren2024pixellm,zhang2024omg,rasheed2024glamm,zhang2024psalm,bao2024cores,xu2023u,wu2025visionllmv2} exploit LLMs to output object hidden states, which are subsequently passed to an additional image-conditioned mask decoder, as shown in \cref{fig:intro_cmp} (b).
Such hidden states usually correspond to some special learnable tokens (\eg , \texttt{<SEG>}~\cite{lai2024lisa}) to adapt to the next-token-prediction paradigm.
However, there are three limitations to this paradigm.
First, due to \textbf{the reliance on powerful image segmentation models}, large language models (LLMs) have insufficient learning of precise spatial localization in images.
Second, there is \textbf{an inconsistency in mask representation between LLM input and output}.
Special tokens serve solely as identifiers and are used as LLM input, resulting in the loss of crucial information from the corresponding hidden states in autoregressive modeling. If users provide mask prompts, they need to extract input prompt features through RoI pooling or cross-attention with visual features~\cite{zhang2024psalm,zhang2024omg}.
Third, \textbf{the overall architecture is complex and heavy}. The mask decoder is typically designed on previous segmentation models and requires the original image to be used again, such as SAM~\cite{kirillov2023sam} and Mask2Former~\cite{cheng2022mask2former}, with some even requiring an additional vision encoder.
Overall, these limitations restrict LMMs from achieving their full potential in image segmentation.

Some early methods view image segmentation as an image generation task~\cite{lu2022unified,bar2022visual}, offering a possible way to address the aforementioned limitations.
By quantizing a mask image into 2D discrete tokens, it is promising to integrate the segmentation mask into LLM input and output sequence consistently.
Although the color space of masks is relatively simple,
it is non-trivial and essential that the output mask \textit{not only captures the object position but also maintains fine-grained shape details}, \eg , object boundary.
VQ-GAN~\cite{esser2021taming} is often used for image quantization that results in a 2D token matrix.
Nonetheless, \textbf{2D token representation is very redundant} for a binarized mask image.
Besides, \textbf{autoregressive patch-wise 2D token prediction has not yet shown very competitive performance} in image generation~\cite{tian2024var}.
These facts pose a challenge for effective mask representation in LMM-based image segmentation.

In this work, we propose \textbf{HiMTok}, an effective \textbf{Hi}erarchical \textbf{M}ask \textbf{Tok}enizer that can represent a mask image using up to 32 hierarchical tokens.
Inspired by TiTok~\cite{yu2024titok}, images can be expressed as highly compact 1D sequences. 
Based on this concept, we introduce a hierarchical mask tokenizer that uses hierarchical mask loss and a causal attention mechanism to \textbf{represent mask images from coarse to fine, which can be learned from abundant and easily accessible mask data}. In this representation, earlier tokens mainly correspond to coarse locations and prototypes, while later tokens focus more on local fine-grained details. 
Each mask token is tightly conditioned on the preceding ones. Hence, this hierarchical design aligns seamlessly with the autoregressive principle of LLMs.
By considering \textbf{mask tokens as a new language}, it enables LMMs to conveniently gain native segmentation capabilities \textbf{without external segmentation foundation models}.
As shown in \cref{fig:intro_cmp} (c), LMM generates discrete mask tokens just as the way for text tokens, and the final segmentation mask is obtained by de-tokenizing the mask tokens.

We devise a three-stage training recipe to progressively integrate image segmentation capabilities into an LMM. 
Initially, our HiMTok is trained on a mask image reconstruction task. 
In the second stage, the LMM and HiMTok are trained together to align visual and language features, utilizing low-resolution images for efficiency.
Finally, the LMM is fine-tuned with high-resolution images to adapt to more general scenarios. Additionally, we introduce a Hierarchical Mask Loss (HML) to facilitate the learning of hierarchical mask tokens, providing explicit supervision across different levels of granularity. We also incorporate a bidirectional information flow between mask tokens and box coordinates within the LMM during training.

Through extensive experiments, our method achieves state-of-the-art performance on image segmentation tasks, including referring expression segmentation, reasoning segmentation and open-vocabulary segmentation.
Our model also improves referring expression comprehension, a visual grounding task. 
These results demonstrate the effectiveness of our elegant HiMTok for multi-task learning in LMM. Meanwhile, our competitive performance in general image understanding tasks indicates that HiMTok enables LMMs to acquire segmentation capabilities without compromising the general abilities.

Succinctly, our contributions are as follows.
\begin{itemize}[leftmargin=0.5cm]
    \item We propose HiMTok, an efficient hierarchical mask tokenizer capable of representing a mask image using up to 32 hierarchical tokens. LMMs are able to learn such token sequences for effective image segmentation, without the need for an image-conditioned mask decoder or off-the-shelf segmentation foundation models.
    \item A three-stage training recipe and a hierarchical mask loss are devised to ensure the progressive learning of HiMTok and LMM. Bidirectional information flow between segmentation and detection is incorporated for the LMM training.
    \item Extensive experiments demonstrate that LMM equipped with HiMTok not only shows superiority in various image segmentation tasks, but also improves visual grounding and maintains the general image understanding capability. Interestingly, visual chain-of-thought by ``outputting mask tokens before box'' improves visual grounding.
\end{itemize}

\section{Related Works}
\label{sec:related}
Large multimodal models (LMMs) have recently become a popular topic.
We have witnessed the continuous improvement in tasks such as visual understanding, OCR, and visual grounding~\cite{wu2024deepseekvl2,chen2024internvl2_5,yao2024minicpm,wang2024qwen2,dubey2024llama3,team2024gemini,li2024aria}.
However, as another fundamental visual task, image segmentation~\cite{zou2023segment} has not been widely included in the capabilities of these open-source large models.
In this section, we will review works on image segmentation paradigms in \cref{fig:intro_cmp} (a) (b) and image tokenization, to better understand our method.

\noindent \textbf{Image segmentation with boundary points.}
Some works regard segmentation masks as polygons, which makes the image segmentation task similar to sequence-to-sequence object detection that generates box coordinates autoregressively~\cite{chen2021pix2seq,wang2022ofa,yang2022unitab,you2023ferret,chen2023shikra}.
Early transformer encoder-decoder models~\cite{zhu2022seqtr,chen2022unified,liu2023polyformer} have proved the feasibility of using boundary points.
LLaFS~\cite{zhu2024llafs} makes use of LLMs in few-shot segmentation, and exploits a refinement network (similar to Mask2Former~\cite{cheng2022mask2former}) after getting a 16-point polygon from LLM.
VistaLLM~\cite{pramanick2024jack} devises an adaptive sampling strategy to better serialize segmentation masks as points for autoregressive mask generation.
This representation struggles with complex shapes and multi-region segmentation.

\noindent \textbf{Image segmentation with LMM hidden states.}
To bring the power of LLMs to image segmentation tasks, LISA~\cite{lai2024lisa} introduces a shared special \texttt{<SEG>} token to abstract the feature of interest that further prompt an additional segmentation foundation model (\eg , SAM~\cite{kirillov2023sam}) to produce the final mask.
GSVA~\cite{xia2024gsva}, LISA++~\cite{yang2023lisapp} and LaSagnA~\cite{wei2024lasagna} expand the number of \texttt{<SEG>} tokens to segment multiple objects.
PixelLM~\cite{ren2024pixellm} incorporates several special tokens to represent multiple granularities and more targets, and designs a lightweight pixel decoder (similar to SAM~\cite{kirillov2023sam}).
PSALM~\cite{zhang2024psalm} incorporates a well-designed input schema and 
uses a powerful segmentation decoder following Mask2Former~\cite{cheng2022mask2former} to improve task generalization.
OMG-LLaVA~\cite{zhang2024omg} integrates OMG-Seg~\cite{li2024omgseg} and an LLM into an end-to-end trainable framework for image understanding and pixel-level reasoning.
These methods manage to adapt the strong reasoning ability of LMMs to the mask decoder, \ie , taking hidden states from LLM as continuous prompts to enable an image-conditioned pixel decoder to segment the objects.

\noindent \textbf{Image tokenization} has been widely explored and applied in visual autoregressive generation.
Many works encode images into 2D discrete token grids~\cite{van2017neural,razavi2019generating,ramesh2021zero,esser2021taming}, which are then flattened into sequences for image generation~\cite{lee2022autoregressive,yu2022scaling}.
VAR~\cite{tian2024var} reformulates visual autoregressive generation as coarse-to-fine next-scale prediction. At each scale, the 2D token grid is predicted simultaneously.
Recently, TiTok~\cite{yu2024titok} manages to tokenize a natural image into a few discrete 1D tokens (\eg 32) in a more compact latent space, which is more efficient.
In our method, we regard a segmentation mask as a special image that is represented as a sequence of 1D coarse-to-fine discrete tokens.



\section{Methodology}

\begin{figure*}[ht]
    \begin{center}
        \includegraphics[width=.97\linewidth]{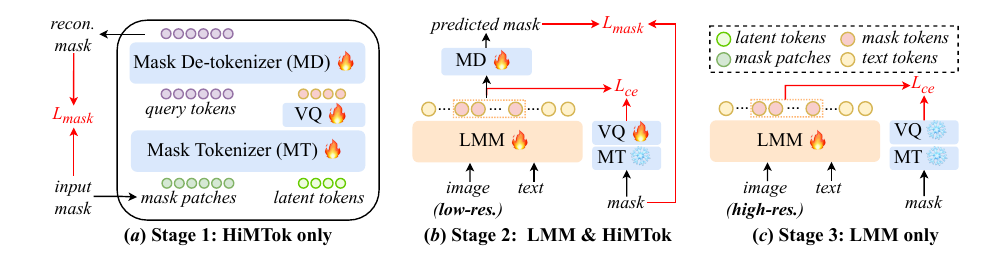}
    \end{center}
    \vspace{-6mm}
    \caption{An overview of HiMTok and the 3-stage training recipe.
    (a) The proposed HiMTok is fully trained in stage 1 by the mask reconstruction task. The VQ loss is emitted here for simplicity. During training, MD takes as input mask tokens of different levels for the hierarchical mask loss.
    (b) The joint training of LMM and parts of HiMTok in stage 2. Images of low-resolution are used as input for efficiency. The cross entropy loss on text tokens is emitted here.
    (c) In stage 3, only the LMM is trained with high-resolution images, which is as simple as common LMM training.
    }
    \label{fig:method}
    \vspace{-4mm}
\end{figure*}

\subsection{Overview}
We propose HiMTok, which uses up to 32 hierarchical 1D tokens to efficiently represent segmentation masks for large language models (LLMs), enabling tokenization and reconstruction of masks.
Together with the lightweight Mask De-tokenizer, LMMs are enabled to elegantly learn object segmentation, with mask tokens functioning like a new language, as shown in \cref{fig:intro_cmp} (c).

To this end, we design a three-stage training recipe for progressive and efficient learning (\cref{fig:method}).
HiMTok is first fully trained to enable the single-modality mask tokenization and de-tokenization.
In stage 2, the joint training of LMM and HiMTok is performed, which brings vision-language alignment in mask tokenization, while still being efficient with images of low resolution.
In stage 3, we focus solely on finetuning the LMM with high-resolution images.

\subsection{Hierarchical Mask Tokenizer}
HiMTok consists of three components: a mask tokenizer (MT), a vector quantization layer (VQ), and a mask de-tokenizer (MD), as illustrated in \cref{fig:method} (a).
Following TiTok~\cite{yu2024titok}, some 1D learnable latent tokens are encoded by MT along with mask image patches.
After quantized by a VQ variant~\cite{zhu2024addressing}, the encoded latent tokens are then reconstructed by MD\footnote{MT annd MD are both standard Transformer layers}.
The number of 1D mask tokens is much smaller than the number of 2D mask patches.

We focus on the hierarchical design here.
Hierarchical tokens are expected to carry coarse-to-fine mask representations, which can be implemented by autoregressive modeling~\cite{tian2024var}.
Considering the assumption of unidirectional dependency, we design a causal attention mechanism for the latent tokens in the attention layers of the mask tokenizer:
the current latent token is conditioned on the mask patches and its former latent tokens.
This mechanism in the mask tokenizer aligns with LLM generation:

\begin{align}
    p\left(m_1,\dots m_K|\mathcal{M}\right)=\prod_{k=1}^K p\left(m_k|\mathcal{M},m_1,\dots m_{k-1}\right),
    \label{eq:mask_tokenizer}
\end{align}
where $\mathcal{M}$ is the input mask, and $m_k$ is the $k$-th mask token, $K$ is the number of mask tokens to represent a mask.
Note that the attention between the input mask patches is bidirectional.

\noindent \textbf{Hierarchical mask loss.}
Explicit supervision on mask tokens of different levels is necessary to ensure hierarchy.
At the $l$-th hierarchical level, the first $l$ mask tokens are de-tokenized to $\hat{M}^{(l)}$ independently.
In the full level, all mask tokens are used.
Based on our observation, few tokens (\eg, 4) usually lead to coarse Gaussian distributions appearing in the mask image after the mask de-tokenizer.
Thus, we empirically exploit the Gaussian blur with different sizes of Gaussian kernel on different levels for multi-grained mask label ($M^{(l)}$) preparation.
Details on how we make coarse mask labels are illustrated in \cref{sup:coarse_mask_labels}.
The hierarchical mask loss (HML) is calculated across the different levels:
\begin{align}
    \mathcal{L}_{mask} = \sum_{l} \mathcal{L}_{mask}^{(l)}\left(\hat{M}^{(l)}, M^{(l)}\right),
\end{align}
where $\mathcal{L}_{mask}^{(l)}$ is the mask loss on the $l$-th level that includes a binary cross entropy loss and a Dice loss~\cite{sudre2017generalised}.
In practice, besides the full level, we sample only a part of other levels following the inverse power-law distribution for the mask loss to train efficiently.

After stage-1 training, HiMTok learns to tokenize segmentation masks into 1D hierarchical tokens in the single mask image modality.


\subsection{Mask-aware Large Multimodal Model}
Regarding the mask tokens in HiMTok as a new language, we can easily incorporate image segmentation capabilities into LMMs.
The only modification required for LMM is an expanded vocabulary with the mask token codebook.

\noindent \textbf{Prompt format.}
To highlight and distinguish the segmentation task and mask tokens from natural language, we exploit some special tokens to indicate them.
Simply put, we use ``\texttt{<ref>referred object</ref>}'' to pinpoint the object to be segmented, and \texttt{<mt\_start>mt\_i $\dots$ <mt\_end>} to indicate the segmentation mask tokens.
Besides, visual grounding is also involved in our experiments as a regular vision-language task, where we use ``\texttt{<box>[x1,y1,x2,y2]</box>}'' to indicate the bounding box.
More details about the prompt design are listed in \cref{sup:prompt_design}.

\noindent \textbf{Bidirectional information flow.}
Detection and segmentation are two classic tasks for object locating.
To learn the inherent relation between masks and bounding boxes, we incorporate bidirectional information flow into our training data, \ie, both box-to-mask and mask-to-box token orders are considered.
LMM is expected to output corresponding mask tokens given a bounding box, and vice versa.
The bounding boxes are generated directly by LMM rather than derived from parsing the de-tokenized masks.
This approach allows the LLM to achieve good consistency and performance in learning detection and segmentation tasks.

\noindent \textbf{Progressive training.}
To enable LMM to learn mask tokens while maintaining its general capability, we train LMM in two stages, as shown in \cref{fig:method} (b) (c).
In stage 2, a massive amount of image segmentation data is used along with some general data.
The mask tokenization is aligned with vision-language through joint training.
Since the LMM encounters various segmentation cases, the understanding of mask tokens is gradually improved.
Low-resolution images are used to train effectively.
Both the cross entropy loss ($L_{ce}$) for next-token-prediction and the hierarchical mask loss are adopted for optimization.
In stage 3, the LMM is further trained with high-resolution images as input.
The amount of segmentation data is reduced to avoid collapse on the general capacity.
Only the cross entropy loss is used.

\noindent \textbf{Inference.}
During inference, if mask tokens are detected in the LLM output, the lightweight mask de-tokenizer will visualize them as the final predicted segmentation mask.
Alternatively, the mask from our de-tokenizer can be further potentially refined by feeding it into an independently finetuned SAM~\cite{kirillov2023sam}. Details on SAM finetuning for optional usage can be found in \cref{sup:finetuning_sam}. We do not utilize SAM in our experiments by default, unless explicitly stated.

\section{Experiments}

\begin{table}[t]
    \centering
    \caption{Training data. $^*$(.) means the dataset is partly sampled. Datasets with $^\#$ are for Mask Perception (\cref{sup:mask_perception}). $\dagger$ means the dataset is synthesized using the open-source engine.}
    \vspace{-3mm}
    \scalebox{0.85}{
        \begin{tabular}{m{.17\columnwidth}<{\centering}|m{.18\columnwidth}<{\centering}|m{.64\columnwidth}<{\centering}}
        \toprule[1.1pt] 
        Key Token & Task & Datasets \\
        \midrule[0.7pt]
        \multirow{8}{*}{\shortstack{Mask\\(2.9M)}} & Sem./Inst. Seg. & \makecell{ADE20K~\cite{zhou2017scene}, PASCAL Context~\cite{mottaghi2014role} \\PartImageNet~\cite{he2022partimagenet}, LVIS-PACO~\cite{ramanathan2023paco}, \\COCO-Rem~\cite{singh2024benchmarking}, COCO-Stuff~\cite{caesar2018coco}\\COCO-Panoptic~\cite{lin2014microsoft}}  \\
        \cline{2-3}
        & Prompt Seg. & SA1B$^*$(1M)~\cite{kirillov2023sam} \\
        \cline{2-3}
        & RES & RefCOCO/+/g~\cite{yu2016modeling,mao2016generation}, gRefCOCO~\cite{liu2023gres}, refCLEF~\cite{kazemzadeh2014referitgame} \\
        \cline{2-3}
        & Reason Seg. & LISA++ Inst. Seg. \& CoT~\cite{yang2023lisapp} \\
        \cline{2-3}
        & Mask Perception & RefCOCO/+/g$^\#$~\cite{yu2016modeling,mao2016generation}, gRefCOCO$^\#$~\cite{liu2023gres} \\
        \hline
        \multirow{4}{*}{\shortstack{Coordinate\\(0.5M)}} & Object Det. & Objects365$^*$(450K)~\cite{shao2019objects365} \\
        \cline{2-3}
        & REC & Cops-Ref~\cite{chen2020cops}, SK-VG~\cite{chen2023advancing}\\
        \cline{2-3}
        & Referential Dialogue & BoxCoT~\cite{chen2023shikra} \\
        \hline
        \multirow{10}{*}{\shortstack{Text\\(3.7M)}} & Caption & InternVL-SA1B-Caption~\cite{chen2024internvl1_5}, ALLaVA-Caption-LAION-4V~\cite{hardy2024allava} \\
        \cline{2-3}
        & VQA & LLaVA-150K~\cite{liu2023llava}, VQAv2~\cite{goyal2017making}, ALLaVA-Instruct~\cite{hardy2024allava}, GQA~\cite{hudson2019gqa}, DOCCI~\cite{onoe2024docci}, CogVLMSFT~\cite{wang2025cogvlm}, SVIT~\cite{zhao2023svit}, SynthClock$\dagger$~\cite{yang2022clock}, AI2D~\cite{kembhavi2016diagram}, MMInstruct$^*$(222K)~\cite{liu2024mminstruct}, Cauldron$^*$(234K)~\cite{laurenccon2024matters} \\
        \cline{2-3}
        & NLP & Evol-Instruct~\cite{hardy2024allava}, Dolly~\cite{conover2023free}, Code-Feedback~\cite{zheng2024opencodeinterpreter}, MathInstruct~\cite{yue2023mammoth}, MetaMathQA~\cite{yu2023metamath}, Orca-Math~\cite{mitra2024orca} \\
        \bottomrule[1.1pt]
        \end{tabular}
    }
    \vspace{-4mm}
    \label{tab:train_data}
\end{table}

\subsection{Training Datasets}
Since the LMM is expected to learn brand-new mask tokens for image segmentation while maintaining its original capability, both segmentation data and general-purpose data are necessary for effective training.
All datasets transcribed in the 3 stages are listed in \cref{tab:train_data}.
For stage-1 training, we exploit the groundtruth masks in all the segmentation-related datasets to train our HiMTok.
Random cropping is adopted to augment the binary mask images.
For stage 2, all the listed datasets are used with a total amount of 7.1 million.
The ratio of segmentation data reaches up to 0.41, which accelerates the learning of mask tokens.
For stage 3, we discard a large number of segmentation and detection samples that are low-quality or overly simplistic in instruction.
The details of the down-sampled datasets are listed in \cref{sup:details}.
As a result, the total amount of training data is decreased to 5.0 million, with the ratio of segmentation data dropping to 0.24, which ensure the general performance.

\subsection{Implementation Details}
In stage 1, HiMTok is initialized with TiTok-L-32~\cite{yu2024titok}.
The resolution of input and reconstructed mask is $256\times 256$.
We use 32 latent tokens to represent a mask image with the codebook size of 1024.
When calculating the hierarchical mask loss, we use 4 levels of mask tokens, including one full-level and three sampled partial levels.
In stage 2, InternVL 2.5~\cite{chen2024internvl2_5} is chosen as the LMM.
The input image size is the base resolution $448\times 448$.
In stage 3, we support high-resolution input images.

\begin{table*}[t]
    \centering
    \caption{Results (cIoU) on the RES benchmarks (RefCOCO/+/g). SFM denotes Segmentation Foundation Models or similar modules, either original or finetuned. ``(ft)'' means the model is finetuned on the in-domain training set.}
    \vspace{-2mm}
    \scalebox{0.95}{
        \begin{tabular}{c|l|c|ccc|ccc|cc}
        \toprule[1.1pt] 
        \multirow{2}{*}{ Paradigm } & \multirow{2}{*}{ Method } & \multirow{2}{*}{ w/ SFM } & \multicolumn{3}{c|}{ RefCOCO } & \multicolumn{3}{c|}{ RefCOCO+ } & \multicolumn{2}{c}{ RefCOCOg }\\
        \cline { 4 - 11 } & & & val & testA & testB & val & testA & testB & val(U) & test(U) \\
        \midrule[0.7pt]
        \multirow{2}{*}{\shortstack{Boundary\\Point-based}} & PolyFormer-B~\cite{liu2023polyformer} & $\times$ & 74.8 & 76.6 & 71.1 & 67.6 & 72.9 & 59.3 & 67.8 & 69.1 \\
        & VistaLLM-7B~\cite{pramanick2024jack} & $\times$ & 74.5 & 76.0 & 72.7 & 69.1 & 73.7 & 64.0 & 69.0 & 70.9 \\
        \hline
        \multirow{11}{*}{\shortstack{Hidden\\State-based}} 
        & LISA-7B(ft)~\cite{lai2024lisa} & $\checkmark$ & 74.9 & 79.1 & 72.3 & 65.1 & 70.8 & 58.1 &  67.9 & 70.6 \\
        & PixelLM-7B~\cite{ren2024pixellm} & $\checkmark$ & 73.0 &76.5 & 68.2 & 66.3 & 71.7 & 58.3 &69.3 & 70.5 \\
        & GSVA-7B~\cite{xia2024gsva} & $\checkmark$ & 76.4 &77.4 &72.8 &64.5 &67.7 & 58.6 & 71.1 & 72.0 \\
        & GSVA-7B(ft)~\cite{xia2024gsva} & $\checkmark$ & 77.2 &78.9 &73.5 &65.9 &69.6 & 59.8 & 72.7 & 73.3 \\
        & LaSagnA-7B~\cite{wei2024lasagna} & $\checkmark$ & 76.8 & 78.7 & 73.8 & 66.4 & 70.6 & 60.1 & 70.6 & 71.9 \\
        & VisionLLM v2~\cite{wu2025visionllmv2} & $\checkmark$ & 76.6 & 79.3 & 74.3 & 64.5 & 69.8 & 61.5 & 70.7 & 71.2 \\
        & OMG-LLaVA ~\cite{zhang2024omg} & $\checkmark$ & 75.6   & 77.7 & 71.2   & 65.6    & 69.7     & 58.9   & 70.7     & 70.2  \\
        & OMG-LLaVA(ft) ~\cite{zhang2024omg} & $\checkmark$ & 78.0   & 80.3  & 74.1   & 69.1    & 73.1     & 63.0   & 72.9     & 72.9  \\
        &GLaMM~\cite{rasheed2024glamm} & $\checkmark$ & 79.5    & 83.2    & 76.9  & 72.6    & 78.7     & 64.6    &  74.2   & 74.9  \\
        & u-LLaVA~\cite{xu2023u} & $\checkmark$  & 83.0   & 85.1    & 80.5  & 77.1  & 81.7  & 70.6    & 77.1  & 78.0 \\
        & PSALM ~\cite{zhang2024psalm} & $\checkmark$ & 83.6   & 84.7  & 81.6   & 72.9    & 75.5     & 70.1   & 73.8     & 74.4 \\ 
        \hline
        \multirow{2}{*}{Others} & GroundHog-7B~\cite{zhang2024groundhog} & $\checkmark$ & 78.5 & 79.9 & 75.7 & 70.5 & 75.0 & 64.9 & 74.1 & 74.6 \\
        & SAM4MLLM-8B~\cite{chen2024sam4mllm} & $\checkmark$ & 79.8 & 82.7 & 74.7 & 74.6 & 80.0 & 67.2 & 75.5 & 76.4 \\
        \hline
        \multirow{3}{*}{\shortstack{Mask\\Token-based}}
        & LMM$_{\text{HiMTok}}$-8B & $\times$ & 81.1 & 81.2 & 79.2 & 77.1 & 78.8 & 71.5 & 75.8 & 76.7 \\
        & LMM$_{\text{HiMTok}}$-8B(ft) & $\times$ & \textbf{85.0} & \textbf{85.2} & \textbf{83.5} & \textbf{79.7} & \textbf{82.7} & \textbf{76.0} & \textbf{80.0} & \textbf{80.6} \\
        \cline{2-11}
        & LMM$_{\text{HiMTok}}$-8B(ft) + SAM & $\checkmark$ & \textbf{85.9} & \textbf{86.3} & \textbf{83.9} & \textbf{80.5} & \textbf{83.7} & \textbf{76.4} & \textbf{80.1} & \textbf{80.9} \\
        \bottomrule[1.1pt]
        \end{tabular}
    }
    \vspace{-4mm}
    \label{tab:res_res}
\end{table*}

\subsection{Referring Expression Segmentation}
Referring Expression Segmentation (RES) is a representative task for evaluation of language-guided segmentation.
We test two versions of our model: one is trained with the 3 stages, the other is further finetuned on the RefCOCO/+/g training set and a small ratio (around 0.3) of mixed general data.
Three classic benchmarks, RefCOCO/+/g~\cite{yu2016modeling,mao2016generation}, are used to evaluate our method.
As shown in \cref{tab:res_res}, LMM equipped with HiMTok (LMM$_{\text{HiMTok}}$-8B) achieves competitive performance without task-specific finetuning and any Segmentation Foundation Models (SFM).
After finetuning, our SFM-free method achieves state-of-the-art performance on all three benchmarks, which not only significantly outperforms previous SFM-free methods, but also beats those with SFM.
Optionally, we obtain further improvement by feeding the mask from our de-tokenizer into a finetuned SAM~\cite{kirillov2023sam}.
The superiority is more significant on RefCOCO+/g where text semantics are more challenging for segmentation, which is owed to the unified modeling of language and segmentation in LMM.

We also evaluate on gRefCOCO~\cite{liu2023gres}, a benchmark for Generalized Referring Expression Segmentation (GRES) that poses challenges in referring to multiple or no objects.
Specifically, we first ask our model whether the object exists.
The segmentation instruction is assigned only when the first response is ``yes''.
\cref{tab:grefcoco} lists the results on gRefCOCO.
Similar conclusions can be observed.
Previous hidden state-based LMMs rely much on SFM, which may restrict the upper bound of their performance on segmenting multiple objects or reject non-existence segmentation.

\subsection{Reasoning Segmentation}
Our method can segment objects given complex and implicit instructions.
Inspired by CoReS~\cite{bao2024cores}, we design a CoT strategy to generate segmentation masks progressively.
Our model is prompted to answer the question with text first and then perform segmentation on the answered objects.

\begin{table}[t]
    \centering
    \caption{Results on generalized referring expression segmentation. * indicates zero-shot performance.}
    \vspace{-2mm}
    \scalebox{0.78}{
        \begin{tabular}{l|cc|cc|cc}
        \toprule[1.1pt] 
        \multirow{2}{*}{ Method } & \multicolumn{2}{c|}{ val } & \multicolumn{2}{c|}{ testA } & \multicolumn{2}{c}{ testB } \\
        \cline { 2 - 7 } & cIoU & gIoU & cIoU & gIoU & cIoU & gIoU \\
        \midrule[0.7pt]
        LISA-7B\cite{lai2024lisa}  & 38.7 & 32.2 & 52.6 & 48.5 & 44.8 &  39.7 \\
        LISA-7B(ft)\cite{lai2024lisa}  & 61.8 & 61.6 & 68.5 & 66.3 & 60.6 & 58.8 \\
        GSVA-7B\cite{xia2024gsva} & 61.7 &63.3 &69.2 &70.1 & 60.3 & 61.3 \\
        GSVA-7B(ft)\cite{xia2024gsva} & 63.3 &66.5 &69.9 &71.1 & 60.5 & 62.2 \\
        LaSagnA*~\cite{wei2024lasagna}  & 38.1 & 32.4 & 50.4 & 47.3 & 42.1 & 38.9 \\
        PSALM*~\cite{zhang2024psalm} & 42.0 &43.3 &52.4 &54.5 & 50.6 & 52.5 \\
        GroundHog-7B~\cite{zhang2024groundhog} & - &66.7 &- &- & - & - \\
        SAM4MLLM-8B~\cite{chen2024sam4mllm} & 67.8 &71.9 &72.2 &\textbf{74.2} & 63.4 & 65.3 \\
        \hline  
        LMM$_{\text{HiMTok}}$-8B & 66.8 & 68.7 & 68.6 & 67.6 & 65.8 & 64.1 \\
        LMM$_{\text{HiMTok}}$-8B(ft) & \textbf{70.4} & \textbf{72.1} & \textbf{74.9} & 73.5 & \textbf{72.0} & \textbf{71.7} \\
        \bottomrule[1.1pt]
        \end{tabular}
    }
    \vspace{-4mm}
    \label{tab:grefcoco}
\end{table}

As shown in \cref{tab:reasonseg}, our method achieves the best performance with the previous SOTA of similar model size in ReasonSeg~\cite{lai2024lisa}, a benchmark featuring complicated and obscure instructions.
Notably, our scores on both the validation and test sets are nearly identical, whereas other methods show lower performance on the test set, which contains a large proportion of short and sophisticated questions.
This further demonstrates that our model not only possesses strong segmentation capabilities but also retains powerful text understanding.

\subsection{Open-vocabulary Segmentation}
Our LMM equipped with HiMTok performs well not only in-domain but also out-of-domain.
We evaluated our model on open-vocabulary segmentation benchmarks, including ADE20K (A-150)~\cite{zhou2019semantic}, PASCAL Context59 (PC-59)~\cite{mottaghi2014role}, and PASCAL VOC 20 (PAS-20)~\cite{everingham2010pascal}.
The large number of defined classes is disadvantageous for our method. Previous segmentation models have typically approached this by either inputting all categories simultaneously for joint modeling or by treating the task as a dense classification problem, resulting in mutual exclusivity between categories. The semantic similarity of some class names can lead to confusion for models that have not been specifically trained to handle such nuances.

\begin{table}[t]
    \centering
    \caption{Results on ReasonSeg.}
    \vspace{-3mm}
    \scalebox{.95}{
        \begin{tabular}{l|cc|cc}
        \toprule[1.1pt] 
        \multirow{2}{*}{ Method } & \multicolumn{2}{c|}{ val } & \multicolumn{2}{c}{ test } \\
        \cline { 2 - 5 } & gIoU & cIoU & gIoU & cIoU \\
        \midrule[0.7pt]
        LISA-7B\cite{lai2024lisa}  & 44.4 & 46.0 & 36.8 & 34.1 \\
        LISA-7B(ft)\cite{lai2024lisa}  & 52.9 & 54.0 & 47.3 & 48.4 \\
        GroundHog-7B~\cite{zhang2024groundhog} & 56.2 &- &- &- \\
        VisionLLM v2~\cite{wu2025visionllmv2}  & 51.0 & - & - & - \\
        LaSagnA~\cite{wei2024lasagna}  & 48.8 & 47.2 & - & - \\
        VISA-7B~\cite{yan2024visa} & 52.7 & 57.8 & - & - \\
        SAM4MLLM-8B~\cite{chen2024sam4mllm} & 58.4 &60.4 &- &- \\
        CoReS-7B~\cite{bao2024cores} & 59.4 & - & 52.4 & - \\
        \hline
        \name -8B & \textbf{60.7} & \textbf{67.0} & \textbf{60.8} & \textbf{66.2} \\
        \bottomrule[1.1pt]
        \end{tabular}
    }
    \vspace{-2mm}
    \label{tab:reasonseg}
\end{table}

We do not feed all the classes into our model to avoid lengthy token sequence and over-finetuning.
Instead, each question contains only one class.
Since many classes do not exist in a single image, we first ask whether the object exists, as we do in gRefCOCO.
Masks of different categories may overlap due to the ambiguity of class names and error from the model.
We empirically prioritize assigning the category of the mask with the smaller area to the confused pixel if several classes are predicted.

As shown in \cref{tab:ov_seg}, our method achieves the best scores in A-150 and PAS-20, while competitive in PC-59.
General language capability of \name  plays a key role in understanding open-vocabulary class names, which is well generalized to the segmentation task.

\begin{table}
    \centering
    \caption{Results (mIoU) on open-vocabulary segmentation.}
    \vspace{-2mm}
    \scalebox{.95}{
        \begin{tabular}{c|ccc}
            \toprule
            Method & A-150 & PC-59 & PAS-20 \\
            \midrule[0.7pt]
            LaSagnA~\cite{wei2024lasagna} & 14.3 & 46.1 & 69.8 \\
            PSALM~\cite{zhang2024psalm} & 18.2 & \textbf{48.5} & 81.3 \\
            \hline
            \name -8B & \textbf{25.0} & 43.9 & \textbf{82.0} \\
            \bottomrule
        \end{tabular}
    }
    \vspace{-4mm}
    \label{tab:ov_seg}
\end{table}

\subsection{Referring Expression Comprehension}
Here we present the results on the REC benchmarks, which are widely recognized for visual grounding evaluation. Our output boxes are generated directly by the LMM following the mask tokens, without any post-processing based on segmentation masks. This approach is highly efficient if our goal is solely object detection, as it bypasses the de-tokenization process.

As illustrated in \cref{tab:rec_bbox}, our method significantly outperforms previous general segmentation models as well as our baseline model, the robust InternVL2.5-8B. This indicates that detection is enhanced through joint training with segmentation. The effect of mask token length on REC is studied in \cref{abl:mask_token_len}.

\subsection{General Image Understanding}
After learning image segmentation, our model continues to maintain its comprehensive image understanding capabilities, which is crucial.

We compare our model with the baseline on MME~\cite{fu2023mme}.
As in \cref{tab:general}, our model is comparable to InternVL2.5-8B across various dimensions.
The joint learning of segmentation even brings some promotion in image understanding of some areas, \eg , position, scene.
However, the performance drops in some dimensions, which is due to the lack of diverse and high-quality data in this segmentation-oriented version.
For example, we do not specially incorporate data on celebrity and OCR.
This also explains why the performance of \name (ft) differs significantly across the dimensions: the segmentation data we used has stronger bias on existence and scene than others.
The model retains intrinsic semantic understanding, despite being fine-tuned exclusively on segmentation data.
More results on the general image understanding are listed in \cref{sup:general}.

\begin{table}[t]
    \centering
    \caption{Results on the REC benchmarks. Acc@0.5 is reported.}
    \vspace{-2mm}
    \scalebox{0.7}{\begin{tabular}{l|ccc|ccc|cc}
    \toprule[1.1pt] 
    \multirow{2}{*}{ Method } &  \multicolumn{3}{c|}{ RefCOCO } & \multicolumn{3}{c|}{ RefCOCO+ } & \multicolumn{2}{c}{ RefCOCOg }\\
    \cline { 2 - 9 } & val & testA & testB & val & testA & testB & val & test \\
    \midrule[0.7pt]
    LISA-7B(ft)~\cite{lai2024lisa}  & 85.4 & 88.8 & 82.6 & 74.2 & 79.5 & 68.4 &  79.3 & 80.4 \\
    GSVA-7B(ft)~\cite{xia2024gsva}  & 86.3 &89.2 &83.8 &72.8 &78.8 & 68.0 & 81.6 & 81.8 \\
    u-LLaVA~\cite{xu2023u}  & 86.0   & 89.5    & 82.3  & 74.1  & 81.2  & 66.7    & 79.9  & 81.7 \\
    \hline
    InternVL2.5-8B   & 90.3   & 94.5    & 85.9  & 85.2  & 91.5  & 78.8    & 86.7  & 87.6 \\
    \hline
    LMM$_{\text{HiMTok}}$-8B & \textbf{92.9} & \textbf{94.7} & \textbf{89.3} & \textbf{87.6} & \textbf{91.5} & \textbf{81.5} & \textbf{88.5} & \textbf{89.0} \\
    \bottomrule[1.1pt]
    \end{tabular}}
    \vspace{-3mm}
    \label{tab:rec_bbox}
\end{table}

\begin{table}[t]
    \centering
    \caption{Results on part of dimensions in MME.}
    \vspace{-2mm}
    \scalebox{0.65}{
        \begin{tabular}{l|ccccccc}
            \toprule[1.1pt] 
            Method & existence & count & position & color & posters & celebrity & scene \\
            \midrule[0.7pt]
            \scalebox{0.95}{InternVL2.5-8B}  & \textbf{200} &  \textbf{170} & 163 & \textbf{180} & \textbf{169} & \textbf{140} & 154  \\
            \hline
            \scalebox{0.95}{LMM$_{\text{HiMTok}}$-8B} & \textbf{200} & 160 & \textbf{166} & \textbf{180} & 164 & 132 & \textbf{163}  \\
            \hline
            \scalebox{0.95}{LMM$_{\text{HiMTok}}$-8B (ft)} & 190 & 113 & 120 & 153 & 57 & 81 & 157  \\
            \bottomrule[1.1pt]
        \end{tabular}
    }
    \vspace{-3mm}
    \label{tab:general}
\end{table}

\subsection{Ablation Study}

\subsubsection{Effect of the mask token length}
\label{abl:mask_token_len}
Mask tokens of different lengths correspond to different granularities for mask representation.
We investigate the effect of mask token length on segmentation (RES) and detection (REC) respectively.

Considering that our model was trained with full-length ($K$ in \cref{eq:mask_tokenizer}) mask token sequences, we finetune it using a mixture of token sequences with varying lengths, by specifying the length in the prompt.
Details on the prompt format can be found in \cref{sup:prompt_design}.
We evaluate on the RefCOCO validation set.

\begin{figure}
    \centering
    \includegraphics[width=1.0\linewidth]{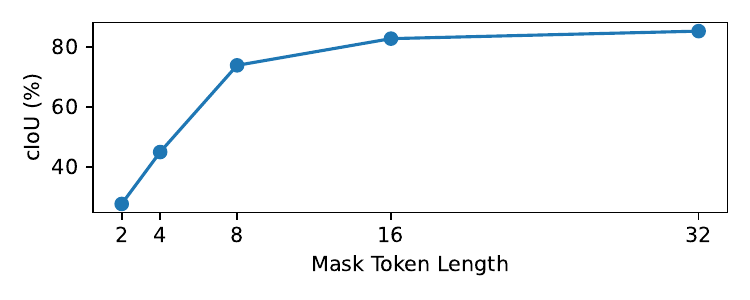}
    \vspace{-8mm}
    \caption{The effect of mask token length on RES. The RefCOCO validation set is used.}
    \label{fig:res_by_token_len}
\end{figure}

\begin{figure}
    \centering
    \includegraphics[width=1.0\linewidth]{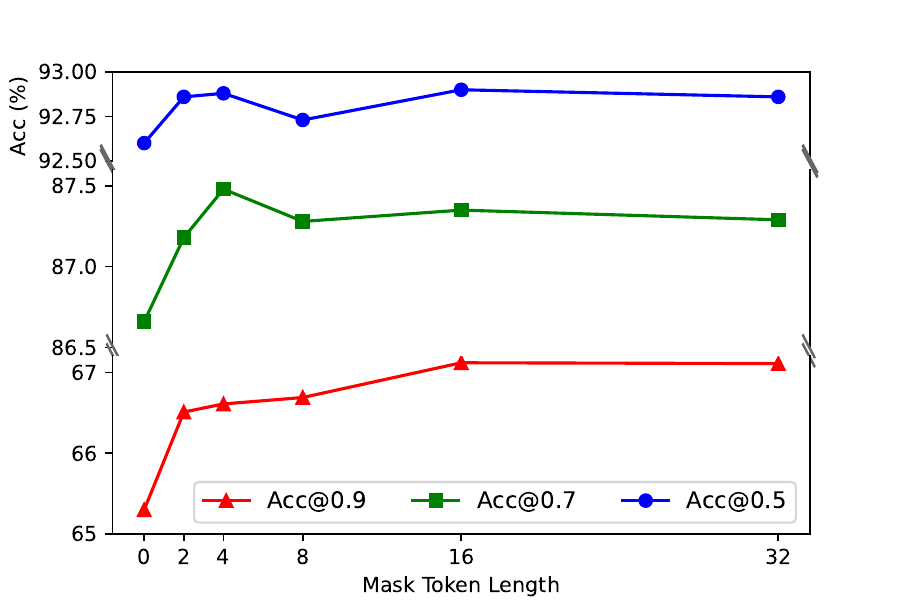}
    \vspace{-5mm}
    \caption{The effect of mask token length on REC with different IoU thresholds. The RefCOCO validation set is used. Note that the acc dimension is not continuous.}
    \vspace{-4mm}
    \label{fig:rec_by_token_len}
\end{figure}

The effect on RES is shown in \cref{fig:res_by_token_len}.
As the mask token length increases, cIoU improves while the gains gradually taper off.
With 16 mask tokens, our method has already achieved 82.8\% cIoU.
By further expanding the token length to 32, we get additional 2.5\% cIoU improvement.
As shown in the right column of \cref{fig:rec_by_token_len}, the fine-grained details become more accurate with longer mask token length.
For simple shapes or less strict scenarios, 16 tokens are enough for representation.

For REC, the object bounding box is predicted after mask tokens of varying lengths.
It is exactly a kind of visual chain-of-thought.
We report accuracy on 3 IoU thresholds: 0.5, 0.7 and 0.9, which correspond to evaluations of varying levels from coarse to fine.
\cref{fig:rec_by_token_len} shows the effect on REC.
As the role segmentation plays, more mask tokens contribute to more refined box prediction.
As more mask tokens are used, acc@0.5 sees a slight improvement, while acc@0.9 improves significantly.
These findings prove the information flow from mask tokens to box coordinates is very useful in both training and evaluation phase.
From some perspective, we can regard the hierarchical mask tokens as a chain-of-thought for visual grounding.
By the way, we do not see improvement on RES for the information flow from box to mask.
This may be due to the fact that the former mask tokens are easier to generate than the direct box coordinates.

\subsubsection{Effect of hierarchical mask loss}
The hierarchical mask loss (HML) is vital for the coarse-to-fine representation learning.
To verify this, we perform 3-stage training without HML.
Instead, the mask loss $\mathcal{L}_{mask}$ consists solely of the full-level mask supervision.

We observe a significant performance degradation without HML.
As shown in \cref{tab:HMT}, RefCOCO+/g, where expression understanding is emphasized, experiences a substantial drop, while RefCOCO shows a relatively minor decline.
We speculate that mask token learning during training without HML compromised the model’s general capability.
Without HML, mask tokens may be learned through shortcuts, and the redundancy within the 32 tokens increases the learning difficulty, potentially disrupting the acquisition and retention of general understanding.

\begin{figure}
    \centering
    \includegraphics[width=.95\linewidth]{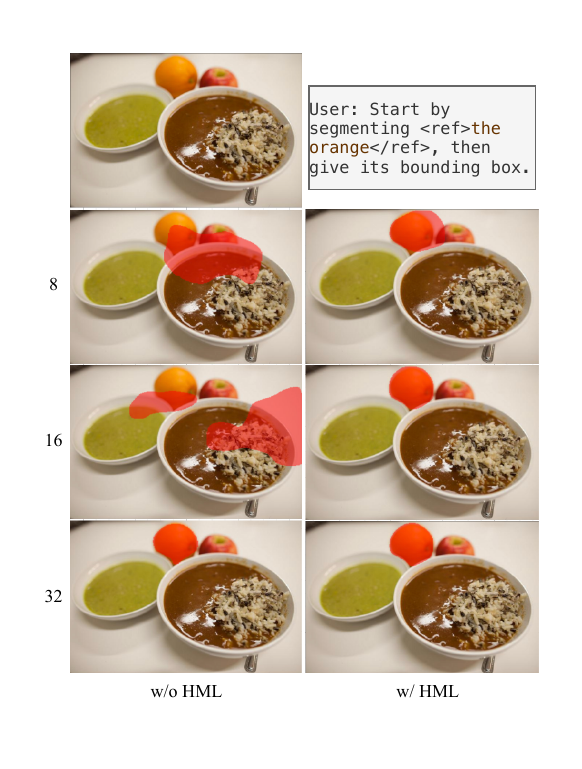}
    \vspace{-2mm}
    \caption{An example that illustrates the effect of different mask token length (8, 16, 32) for our model trained with and without HML. For model without using HML, full tokens are restrictly required for the mask quality, while model with HML can segment simple objects with fewer tokens.}
    \vspace{-1mm}
    \label{fig:cases-hml}
\end{figure}

As shown on the left of \cref{fig:cases-hml}, mask tokens fewer than 32 lead to completely incorrect segmentation masks, which suggests that training without HML strictly requires the use of full-length mask tokens throughout.
In contrast, the use of our hierarchical mask tokens is flexible.
More interesting visualization cases are presented in \cref{sup:cases_more}.

\begin{table}
    \centering
    \caption{Ablation on the hierarchical mask loss (HML). The validation set is used for comparison.}
    \vspace{-3mm}
    \begin{tabular}{c|ccc}
        \toprule
        HML & RefCOCO & RefCOCO+ & RefCOCOg \\
        \midrule[0.7pt]
        $\times$ & 79.2 & 64.7 & 63.9 \\
        $\checkmark$ & \textbf{81.1} & \textbf{77.1} & \textbf{75.8} \\
        \bottomrule
    \end{tabular}
    \vspace{-3mm}
    \label{tab:HMT}
\end{table}



\section{Conclusion}
We present HiMTok, a Hierarchical Mask Tokenizer that represents segmentation masks as hierarchical mask token sequences. Existing LMMs equipped with HiMTok are capable of learning to segment objects specified by language prompts. Hierarchical mask loss is proposed to ensure the learning of coarse-to-fine mask tokens. We also develop a 3-stage training recipe for progressive learning of segmentation and general visual capabilities. Extensive experiments demonstrate the effectiveness of HiMTok in a variety of visual and segmentation tasks.


{
    \small
    \bibliographystyle{ieeenat_fullname}
    \bibliography{main}
}

\clearpage
\setcounter{page}{1}
\maketitlesupplementary
\appendix



\section{Limitations}
While HiMTok enables large multimodal models to acquire native (referring) image segmentation capabilities in a concise and natural manner, the current work has some limitations.
(1) The length of predicted mask tokens is pre-defined. LMMs are not able to determine it adaptively according to object shape complexities.
(2) The current model is relatively passive for object segmentation, due to the use of passive segmentation training data.
We need to specify referring expressions to clarify the expected objects, rather than let the model itself segment all objects of interest at once.
(3) It appears challenging for fine-grained region segmentation in the current version (See \cref{res_fine}). The lack of multi-scale feature design may cause the loss of fine-grained features.

\section{Multi-grained mask labels}
\label{sup:coarse_mask_labels}
The multi-grained mask labels are important in the hierarchical mask loss.
We have tried to use only the final mask label to supervise each granularity level, but the loss is relatively high due to the insufficient representation by few mask tokens.
This also causes unstable training and even side influence on the general capability of LMM.
In experiments, we find that few mask tokens are usually de-tokenized into Gaussian distribution maps, which inspire us to make multi-grained mask labels by Gaussian blurring.

\begin{figure*}[ht]
    \begin{center}
        \includegraphics[width=1.0\linewidth]{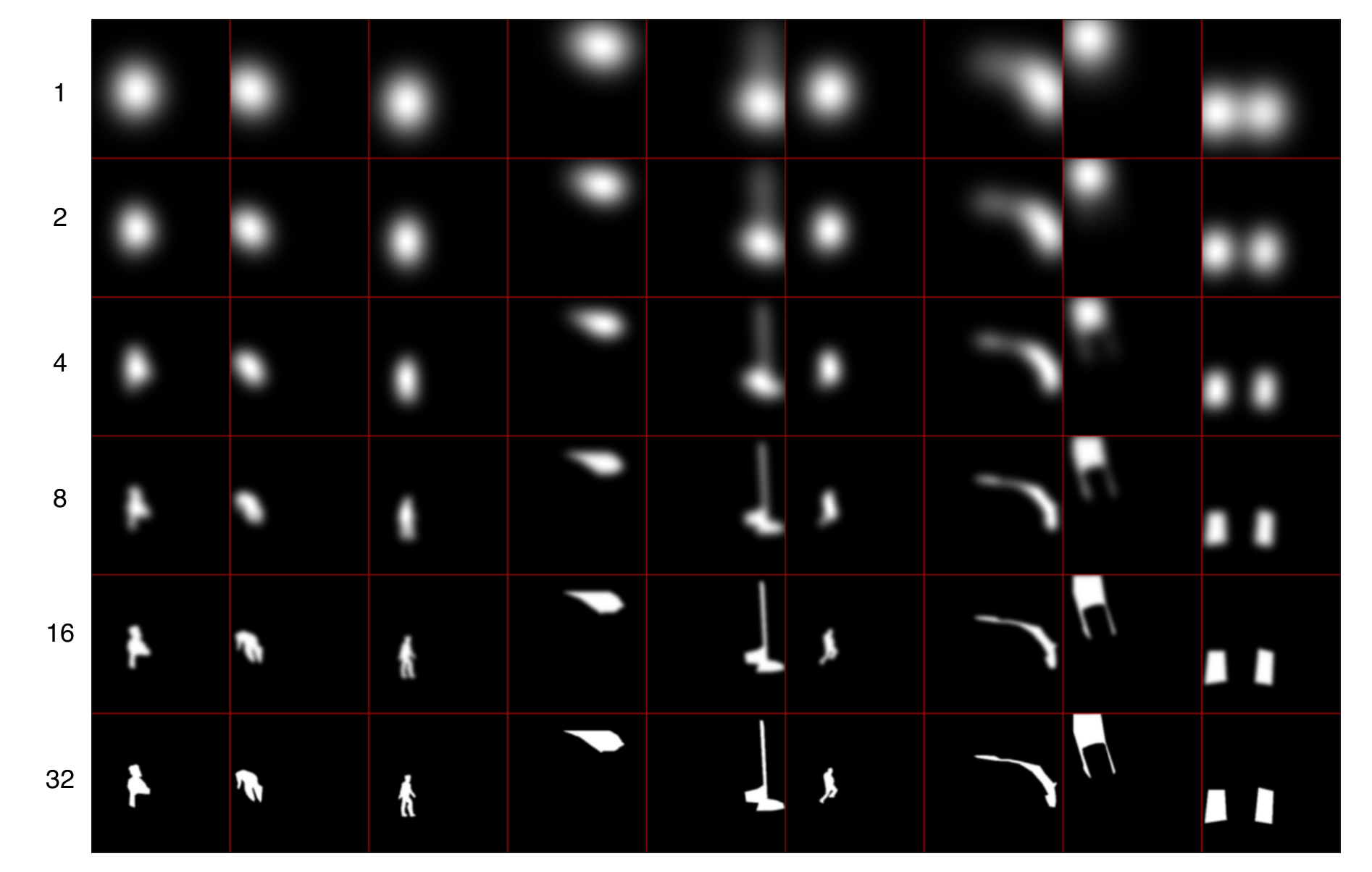}
    \end{center}
    \caption{The Gaussian-blurred mask label at different levels.}
    \label{fig:gaussian_blur}
\end{figure*}

Under the setting of mask token length to 32, we first choose a full-level (\ie , 32 tokens) sequence and 3 random levels that are sampled by $p_l=\frac{1}{l+8},1\leq l < 32$.
For each partial level $l$, we produce a kernel following the 2D Gaussian distribution function $\mathcal{N}(\mu, \sigma)$, where $\mu=0$, and $\sigma =\frac{100}{l+1}-2$. 
Then the kernel is applied to the full-level mask image to obtain the mask label at level $l$.
\cref{fig:gaussian_blur} visualizes some examples with different granularity levels.

\section{Finetuning SAM}
\label{sup:finetuning_sam}
\begin{figure}[t]
    \centering
    \includegraphics[width=1.0\linewidth]{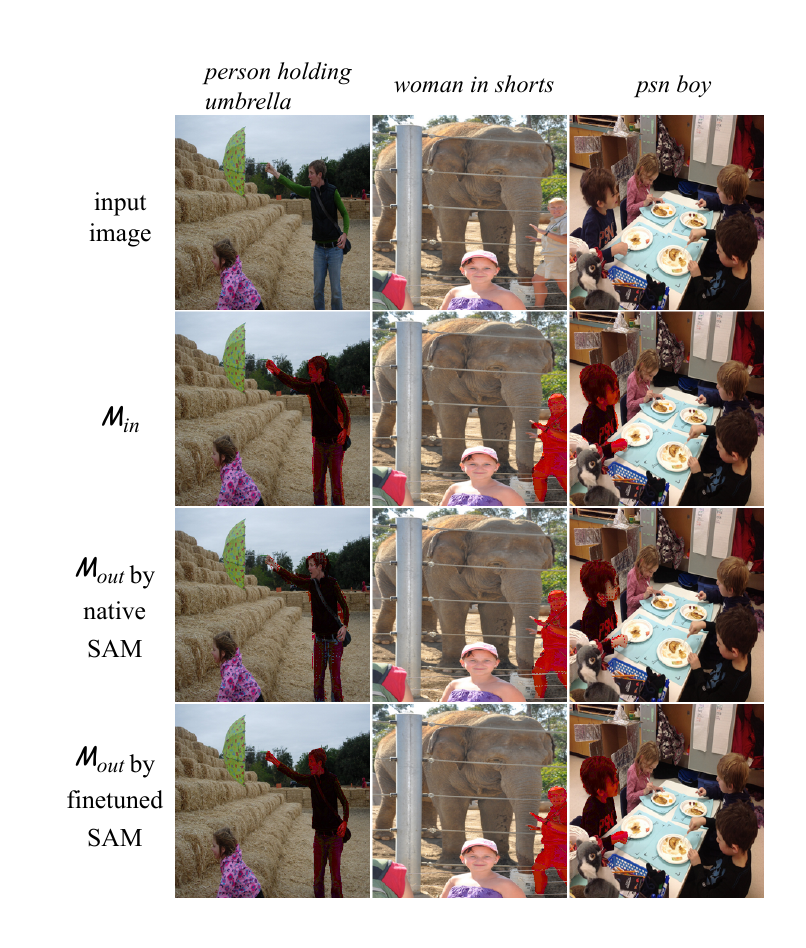}
    \caption{Further improvements by finetuned SAM.}
    \label{fig:sam_effect}
    \vspace{-0.5em}
\end{figure}

Despite the fact that LMM equipped with HiMTok, without relying on segmentation foundation models, has achieved state-of-the-art performance on various segmentation tasks, further improvements can still be expected by feeding the mask from our de-tokenizer into SAM~\cite{kirillov2023sam}.
This post-refinement module can be defined as a mapping $\mathcal{R}:(\mathcal{I},\mathcal{M}_{in})\to \mathcal{M}_{out}$, where $\mathcal{M}_{in}$ is the output mask by our HiMTok-equipped LMM, and $\mathcal{I}$ is the input image.
However, we find that the native SAM given $\mathcal{M}_{in}$ tends to segment fine-grained parts of objects or generate mask maps with holes, even when the given mask clearly covers large and unambiguous regions, as visualized in \cref{fig:sam_effect}.

To adapt to our case, we finetune the mask decoder and the mask convolution layers in SAM.
The input mask $\mathcal{M}_{in}$ is augmented in two aspects.
On the one hand, the number of mask tokens passed to the de-tokenizer is randomly reduced so that the fine-grained shape details may be lost.
On the other hand, the de-tokenized mask is processed by random morphological augmentation, including dilation and erosion.
As a result, the finetuned SAM is able to refine imperfect masks.
Qualitative results are shown in \cref{fig:sam_effect}.
The finetuned SAM improve the edge details of the segmentation masks.

\section{More details on training}
\label{sup:details}
In stage-3, some datasets are down-sampled.
Details are listed in \cref{tab:num_downsampled}.
The initial learning rate is $4e-5$ in stage 2, $2e-5$ in stage 3, and $8e-6$ in task finetuning.
The GPU hours (Nvidia A800) for our 3 stages are: 192, 1920 and 640.

\begin{table}
    \centering
    \caption{The amount of samples in down-sampled datasets for stage-3.}
    \begin{tabular}{c|cc}
        \toprule
        Dataset & stage-2 & stage-3 \\
        \hline
        SA1B & 1M & 250K \\
        COCO-Rem & 350K & 35K \\
        COCO-Stuff & 500K & 50K \\
        COCO-Panoptic & 233K & 116K \\
        PartImageNet & 20K & 4K \\
        Objects365 & 450K & 90K \\
        \bottomrule
    \end{tabular}
    \label{tab:num_downsampled}
\end{table}

\section{Mask Perception}
\label{sup:mask_perception}
One key feature of our method is the consistence of mask representation in LLM input and output.
The image segmentation tasks are all text-to-mask.
As a supplement, we devise a mask-to-text task, Mask Perception\footnote{https://huggingface.co/datasets/yayafengzi/Mask\_Perception}, which aims for fine-grained understanding.
Mask Perception (MaP) mirrors RES: models are required to choose a matched expression, given an image, an object mask and several expression options.

The training and test sets are built based on RefCOCO/+/g~\cite{yu2016modeling,mao2016generation}.
We randomly sample some examples and reverse the positions of masks and expressions in multiple-choice questions.
One or more positive options are possible in the training set, while only one positive option is available in the test set to simplify evaluation. 
Negative expression options are selected from other objects in the same image or from different images. This approach encourages the model to distinguish the distinctive features of various parts.
Statistics are shown in \cref{tab:statistics_mask_perception}.

We have tried to compare our method with PSALM~\cite{zhang2024psalm} which performs interactive segmentation well.
However, we found that PSALM does not follow instructions well in our MaP test set.
Therefore, we compared our method both with and without the MaP training set.
As shown in \cref{tab:res_mask_perception}, our method inherently has a good perception of input masks. With the addition of the MaP training set, the mask perception capability is significantly improved.

\begin{table}
    \centering
    \caption{Statistics of Mask Perception. The data source is RefCOCO/+/g.}
    \label{tab:statistics_mask_perception}
    \begin{tabular}{c|ccc}
        \toprule
        data split & source & single/multiple choices & No. \\
        \hline
        Training &  train & multiple & 190k \\
        Test & val \& test & single & 10k \\
        \bottomrule
    \end{tabular}
\end{table}

\begin{table}
    \centering
    \caption{Accuracy on mask perception.}
    \label{tab:res_mask_perception}
    \begin{tabular}{c|cc}
        \toprule
        w/ MaP & $\times$ & $\checkmark$ \\
        \hline
        Acc &  63.1 & 81.8 \\
        \bottomrule
    \end{tabular}
\end{table}


\section{Results on fine-grained regions}
\label{res_fine}
Given that HiMTok decodes masks directly via mask tokens, a natural concern is whether it can maintain high segmentation quality for fine-grained regions without leveraging the fine-grained features of the original image.

Here we show the results on small object segmentation in RefCOCO/+/g. Objects whose mask areas occupy less than 4\% of the image are considered as small, resulting in 12.8\% samples.
Shown in \cref{tab:results_small}, the cIoUs of ours still have significant priority compared to PSALM~\cite{zhang2024psalm}.
However, the cIoU scores fall significantly behind the overall performance (\cref{tab:res_res}), which highlights a common challenge.

\begin{table*}[h]
    \centering
    \vspace{-4mm}
    \caption{Results on small object segmentation.}
    \label{tab:results_small}
    \vspace{-4mm}
    \scalebox{1.0}{
        \begin{tabular}{l *{8}{c}}
        \toprule
        & \multicolumn{3}{c}{Ref COCO} & \multicolumn{3}{c}{Ref COCO+} & \multicolumn{2}{c}{Ref COCOg} \\
        \cmidrule(r){2-4} \cmidrule(r){5-7} \cmidrule(r){8-9}
        & val & testA & testB & val & testA & testB & val & test \\
        \midrule
        PSALM~\cite{zhang2024psalm} & 64.50 & 68.43 & 57.31 & 45.58 & 55.89 & 38.14 & 51.10 & 48.78 \\
        ours & 67.15 & 75.00 & 60.34 & 56.81 & 63.06 & 48.29 & 57.71 & 55.99 \\
        \bottomrule
        \end{tabular}
    }
\end{table*}

We also review the segmentation boundaries.
Bfscore (a boundary-aware F1 metric) on RefCOCO (val) is reported:
our method gets 0.927, which is competitive to PSALM (0.936).
PSALM integrates Mask2Former that is favored by multi-scale features.
We believe there are rooms for future exploration in our paradigm.

\section{Results on general image understanding}
\label{sup:general}
Here, we list additional results on general image understanding. Our model is not finetuned on these tasks.

We compare methods on MME~\cite{fu2023mme} Perception, VQAv2~\cite{goyal2017making}, and POPE~\cite{li2023evaluating}. As shown in \cref{tab:general_suppl}, our method is comparable to state-of-the-art LMMs, except in the areas of landmarks, artwork, and OCR, where we do not specially utilize corresponding training data in this work.
We can conclude that \name is a comprehensional and general LMM.

\begin{table*}[t]
    \centering
    \caption{Results on general image understanding.}
    \scalebox{0.8}{
        \begin{tabular}{l|cccccccccc|c|c}
            \toprule[1.1pt] 
            \multirow{2}{*}{Method} & \multicolumn{10}{c|}{MME} & \multirow{2}{*}{VQAv2} & \multirow{2}{*}{POPE}  \\
            \cline{2-11}
            & existence & count & position & color & posters & celebrity & scene & landmark & artwork & OCR & & \\
            \midrule[0.7pt]
            PSALM~\cite{zhang2024psalm}  & - & - & - & - & - & - & - & - & - & - & 62.3 & 80.3 \\
            InternVL2.5-8B~\cite{chen2024internvl2_5}  & \textbf{200} &  \textbf{170} & 163 & \textbf{180} & \textbf{169} & \textbf{140} & 155 & \textbf{172} & \textbf{160} & \textbf{178} & -    & \textbf{90.6}  \\
            \hline
            LMM$_{\text{HiMTok}}$-8B & \textbf{200} & 160 & \textbf{166} & \textbf{180} & 164 & 132 & \textbf{163} & 132 & 120 & 88 & \textbf{75.9} & 86.8  \\

            \bottomrule[1.1pt]
        \end{tabular}
    }
    \label{tab:general_suppl}
\end{table*}

\section{More visualizations}
\label{sup:cases_more}
\begin{figure}[t]
    \centering
    \includegraphics[width=1.0\linewidth]{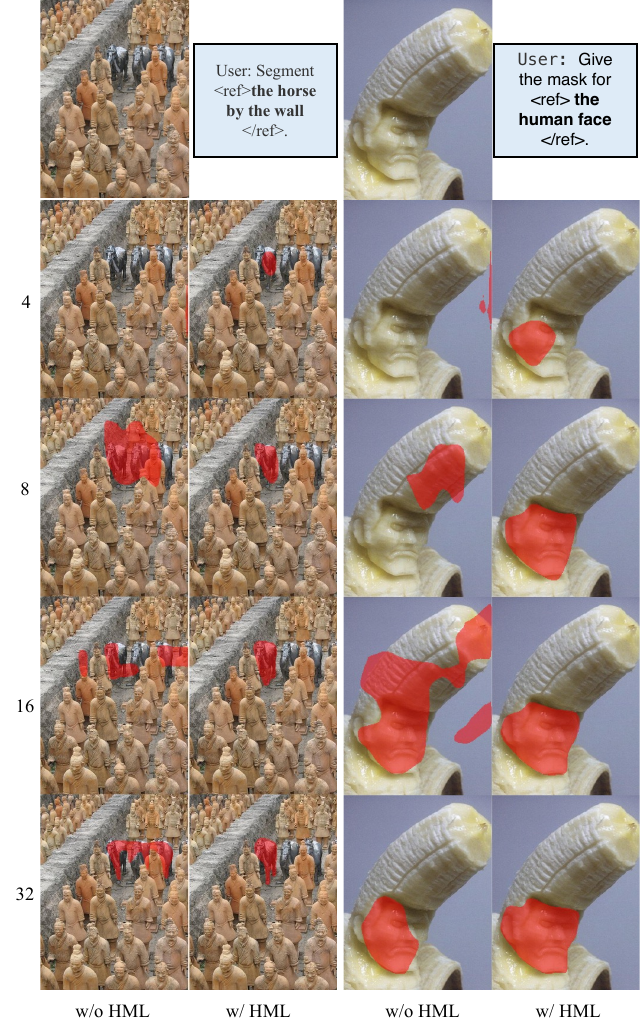}
    \caption{More examples on the coarse-to-fine mask token representation with and without HML.}
    \label{fig:cases_hml_more}
    \vspace{-0.5em}
\end{figure}

\begin{figure}[t]
    \centering
    \includegraphics[width=1.0\linewidth]{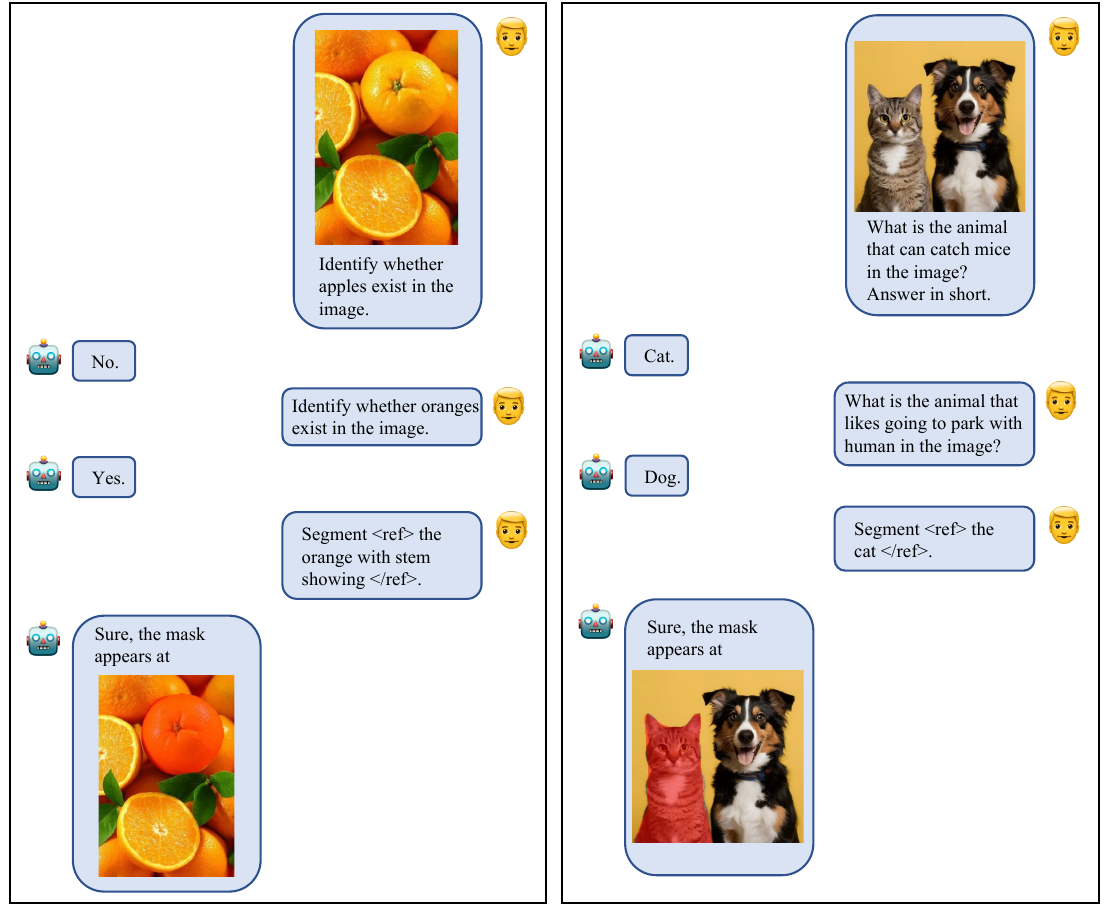}
    \caption{Referring image segmentation in conversation.}
    \label{fig:conv_seg}
    \vspace{-0.5em}
\end{figure}

\cref{fig:cases_hml_more} shows more interesting and challenging cases.
\cref{fig:conv_seg} illustrates how we implement referring image segmentation in conversations.
This is what we do when evaluating our model on gRefCOCO, ReasonSeg, and open-vocabulary segmentation.

\section{Prompt design}
\label{sup:prompt_design}
We prepared plentiful prompt templates for instruction tuning on segmentation and visual grounding.

For the bidirectional information flow between segmentation and grounding, \cref{tab:mask_then_box_questions,tab:mask_then_box_answers} list templates for mask-to-box, and \cref{tab:box2mask_questions,tab:points2mask_questions} are for coordinate-to-mask.
\cref{tab:mask_answers} list templates for segmentation-only responses.
Templates in \cref{tab:abl_token_len_questions} are used to specify the mask token length from LMM. 
If visual grounding is the only target without mask tokens, we can refer to \cref{tab:abl_det}.

\begin{table}[ht]
    \centering
    \caption{Templates of instruction for segmentation then grounding.}
    \begin{minipage}{1.0\columnwidth}\vspace{0mm}
        \centering
        \begin{tcolorbox}
            \centering
            \scriptsize
            \hspace{-6mm}
            \begin{itemize}[leftmargin=0mm]
                \setlength{\itemsep}{2pt}
                \item ``First create the segmentation mask for \textless ref\textgreater\{\}\textless /ref\textgreater, then identify its bounding box."
                \item ``Begin by generating the mask for \textless ref\textgreater\{\}\textless /ref\textgreater, followed by determining its box coordinates."
                \item ``Start with the detailed mask of \textless ref\textgreater\{\}\textless /ref\textgreater, then locate its containing box."
                \item ``Initially segment \textless ref\textgreater\{\}\textless /ref\textgreater, then find its bounding box region."
                \item ``First produce the mask for \textless ref\textgreater\{\}\textless /ref\textgreater, then specify its box location."
                \item ``Commence with the mask of \textless ref\textgreater\{\}\textless /ref\textgreater, then determine its box boundaries."
                \item ``Start by segmenting \textless ref\textgreater\{\}\textless /ref\textgreater, then outline its bounding region."
                \item ``First extract the mask of \textless ref\textgreater\{\}\textless /ref\textgreater, then mark its box coordinates."
                \item ``Begin with the precise mask for \textless ref\textgreater\{\}\textless /ref\textgreater, then identify its box location."
                \item ``Initially create the mask for \textless ref\textgreater\{\}\textless /ref\textgreater, then define its bounding box."
            \end{itemize}
        \end{tcolorbox}
    \vspace{-2mm}
    \end{minipage}
    \label{tab:mask_then_box_questions}
\end{table}

\begin{table}[ht]
    \centering
    \caption{Templates of response for segmentation then grounding.}
    \begin{minipage}{0.99\columnwidth}\vspace{0mm}
        \centering
        \begin{tcolorbox} 
            \centering
            \scriptsize
            \hspace{-6mm}
            \begin{itemize}[leftmargin=0mm]
                \setlength{\itemsep}{2pt}
                \item ``Certainly, you can find the mask at \{\}, and the box is represented as \textless box\textgreater\{\}\textless /box\textgreater."
                \item ``Of course, the mask located is \{\}, while the box is shown as \textless box\textgreater\{\}\textless /box\textgreater."
                \item ``Absolutely, the mask is situated at \{\}, with the box described as \textless box\textgreater\{\}\textless /box\textgreater."
                \item ``Sure, the mask appears at \{\}, and here's the box: \textless box\textgreater\{\}\textless /box\textgreater."
                \item ``Indeed, the mask can be found at \{\}, with the box marked as \textless box\textgreater\{\}\textless /box\textgreater."
                \item ``OK, the mask is at \{\}, and the box is indicated as \textless box\textgreater\{\}\textless /box\textgreater."
                \item ``Affirmative, you have the mask at \{\}, and the box is designated \textless box\textgreater\{\}\textless /box\textgreater."
                \item ``Got it, the mask is at \{\}, and you will see the box like \textless box\textgreater\{\}\textless /box\textgreater."
                \item ``Sure, the mask appears at \{\}, and here is the box represented: \textless box\textgreater\{\}\textless /box\textgreater."
                \item ``Yes indeed, find the mask at \{\}, and the box outlined as \textless box\textgreater\{\}\textless /box\textgreater."
            \end{itemize}
        \end{tcolorbox}
        \vspace{-2mm}
    \end{minipage}
    \label{tab:mask_then_box_answers}
\end{table}

\begin{table}[ht]
    \centering
    \caption{Templates of instruction for box/point-prompted segmentation in SA1B.}
    \begin{minipage}{0.99\columnwidth}\vspace{0mm}
        \centering
        \begin{tcolorbox} 
            \centering
            \scriptsize
            \hspace{-6mm}
            \begin{itemize}[leftmargin=0mm]
            \setlength{\itemsep}{2pt}
                \item ``Generate the segmentation mask for the object located within the bounding box \textless box\textgreater\{\}\textless /box\textgreater and at the points \{\} in the provided image."
                \item ``Extract the mask for the object inside the \textless box\textgreater\{\}\textless /box\textgreater bounding box and corresponding to the points \{\} in the given image."
                \item ``Create the mask for the object found in the region defined by \textless box\textgreater\{\}\textless /box\textgreater and identified by the points \{\} in this image."
                \item ``Identify the object mask within the bounding box \textless box\textgreater\{\}\textless /box\textgreater and marked by the points \{\} in the image."
                \item ``Acquire the segmentation mask of the object enclosed by \textless box\textgreater\{\}\textless /box\textgreater and situated at the points \{\} in the image."
                \item ``Determine the mask for the object that is located within the area specified by \textless box\textgreater\{\}\textless /box\textgreater and identified by the points \{\} in the image."
                \item ``Produce the mask for the object situated in the \textless box\textgreater\{\}\textless /box\textgreater box and at the points \{\} within the image."
                \item ``Locate the mask for the object that lies inside the \textless box\textgreater\{\}\textless /box\textgreater boundaries and corresponds to the points \{\} in the image."
                \item ``Segment the mask for the object found within the bounds of \textless box\textgreater\{\}\textless /box\textgreater and marked at the points \{\} in the given image."
                \item ``Outline the mask of the object residing within the box \textless box\textgreater\{\}\textless /box\textgreater and corresponding to the points \{\} in the image."
            \end{itemize}
        \end{tcolorbox}
        \vspace{-2mm}
    \end{minipage}
    \label{tab:box2mask_questions}
\end{table}

\begin{table}[ht]
    \centering
    \caption{Templates of instruction for point-prompted segmentation in SA1B.}
    \begin{minipage}{0.99\columnwidth}\vspace{0mm}
        \centering
        \begin{tcolorbox} 
            \centering
            \scriptsize
            \hspace{-6mm}
            \begin{itemize}[leftmargin=0mm]
            \setlength{\itemsep}{2pt}
                \item ``Generate the segmentation mask for the object located at the points \{\} in the provided image."
                \item ``Extract the mask for the object that corresponds to the points \{\} in the given image."
                \item ``Create the mask for the object identified by the points \{\} in this image."
                \item ``Identify the object mask corresponding to the points \{\} in the image."
                \item ``Acquire the segmentation mask of the object marked at the points \{\} in the image."
                \item ``Determine the mask for the object located at the specified points \{\} in the image."
                \item ``Produce the mask for the object situated at the points \{\} within the image."
                \item ``Locate the mask for the object that is identified by the points \{\} in the image."
                \item ``Segment the mask for the object found at the points \{\} in the given image."
                \item ``Outline the mask of the object located at the specified points \{\} in the image."
            \end{itemize}
        \end{tcolorbox}
        \vspace{-2mm}
    \end{minipage}
    \label{tab:points2mask_questions}
\end{table}

\begin{table}[ht]
    \centering
    \caption{Templates of response for segmentation only.}
    \begin{minipage}{0.99\columnwidth}\vspace{0mm}
        \centering
        \begin{tcolorbox} 
            \centering
            \scriptsize
            \hspace{-6mm}
            \begin{itemize}[leftmargin=0mm]
            \setlength{\itemsep}{2pt}
                \item ``Certainly, the mask is located at \{\}."
                \item ``Of course, you can find the mask at \{\}."
                \item ``Absolutely, the mask is available at \{\}."
                \item ``Sure, the mask is positioned at \{\}."
                \item ``Indeed, the mask appears at \{\}."
                \item ``OK, the mask is situated at \{\}."
                \item ``Affirmative, the mask can be found at \{\}."
                \item ``Got it, the mask is at \{\}."
                \item ``Sure, the mask is present at \{\}."
                \item ``Yes indeed, you will find the mask at \{\}."
            \end{itemize}
        \end{tcolorbox}
        \vspace{-2mm}
    \end{minipage}
    \label{tab:mask_answers}
\end{table}

\begin{table}[ht]
    \centering
    \caption{Templates of instruction for segmentation with specified token lengths.}
    \begin{minipage}{0.99\columnwidth}\vspace{0mm}
        \centering
        \begin{tcolorbox} 
            \centering
            \scriptsize
            \hspace{-6mm}
            \begin{itemize}[leftmargin=0mm]
                \setlength{\itemsep}{2pt}
                \item ``First create the segmentation mask for \textless ref\textgreater\{\}\{\}\textless /ref\textgreater\{\} by \{\} tokens, then identify its bounding box."
                \item ``Begin by generating the mask for \textless ref\textgreater\{\}\{\}\textless /ref\textgreater\{\} by \{\} tokens, followed by determining its box coordinates."
                \item ``Start with the detailed mask of \textless ref\textgreater\{\}\{\}\textless /ref\textgreater\{\} by \{\} tokens, then locate its containing box."
                \item ``Initially segment \textless ref\textgreater\{\}\{\}\textless /ref\textgreater\{\} by \{\} tokens, then find its bounding box region."
                \item ``First produce the mask for \textless ref\textgreater\{\}\{\}\textless /ref\textgreater\{\} by \{\} tokens, then specify its box location."
                \item ``Commence with the mask of \textless ref\textgreater\{\}\{\}\textless /ref\textgreater\{\} by \{\} tokens, then determine its box boundaries."
                \item ``Start by segmenting \textless ref\textgreater\{\}\{\}\textless /ref\textgreater\{\} by \{\} tokens, then outline its bounding region."
                \item ``First extract the mask of \textless ref\textgreater\{\}\{\}\textless /ref\textgreater\{\} by \{\} tokens, then mark its box coordinates."
                \item ``Begin with the precise mask for \textless ref\textgreater\{\}\{\}\textless /ref\textgreater\{\} by \{\} tokens, then identify its box location."
                \item ``Initially create the mask for \textless ref\textgreater\{\}\{\}\textless /ref\textgreater\{\} by \{\} tokens, then define its bounding box."
            \end{itemize}
        \end{tcolorbox}
        \vspace{-2mm}
    \end{minipage}
    \label{tab:abl_token_len_questions}
\end{table}

\begin{table}[ht]
    \centering
    \caption{Templates of for visual grounding without mask tokens.}
    \begin{minipage}{0.99\columnwidth}\vspace{0mm}
        \centering
        \begin{tcolorbox} 
            \centering
            \small
            Instructions
            \scriptsize
            \hspace{-6mm}
            \begin{itemize}[leftmargin=0mm]
                \setlength{\itemsep}{2pt}
                \item ``Please provide the bounding box of \textless ref\textgreater{}\{\}\textless /ref\textgreater{}."
                \item ``Locate the bounding box for \textless ref\textgreater{}\{\}\textless /ref\textgreater{}."
                \item ``Identify the box coordinates of \textless ref\textgreater{}\{\}\textless /ref\textgreater{}."
                \item ``Determine the bounding region for \textless ref\textgreater{}\{\}\textless /ref\textgreater{}."
                \item ``Find the box boundaries of \textless ref\textgreater{}\{\}\textless /ref\textgreater{}."
                \item ``Mark the box location of \textless ref\textgreater{}\{\}\textless /ref\textgreater{}."
                \item ``Specify the box coordinates for \textless ref\textgreater{}\{\}\textless /ref\textgreater{}."
                \item ``Outline the bounding box of \textless ref\textgreater{}\{\}\textless /ref\textgreater{}."
                \item ``Can you show the box for \textless ref\textgreater{}\{\}\textless /ref\textgreater{}?"
            \end{itemize}
            \vspace{2mm}
            \small
            Responses
            \scriptsize
            \hspace{-6mm}
            \begin{itemize}[leftmargin=0mm]
                \setlength{\itemsep}{2pt}
                \item ``Certainly, the box is located at \textless box\textgreater{}\{\}\textless /box\textgreater{}."
                \item ``Of course, here's the bounding box: \textless box\textgreater{}\{\}\textless /box\textgreater{}."
                \item ``The requested box coordinates are \textless box\textgreater{}\{\}\textless /box\textgreater{}."
                \item ``Found the bounding region: \textless box\textgreater{}\{\}\textless /box\textgreater{}."
                \item ``Here's the box location: \textless box\textgreater{}\{\}\textless /box\textgreater{}."
                \item ``The object's box is marked as \textless box\textgreater{}\{\}\textless /box\textgreater{}."
                \item ``Bounding box determined: \textless box\textgreater{}\{\}\textless /box\textgreater{}."
                \item ``Located the boundaries at \textless box\textgreater{}\{\}\textless /box\textgreater{}."
            \end{itemize}
        \end{tcolorbox}
        \vspace{-2mm}
    \end{minipage}
    \label{tab:abl_det}
\end{table}

\end{document}